\documentclass{article}

\usepackage{PRIMEarxiv}

\usepackage[utf8]{inputenc} 
\usepackage[T1]{fontenc}    
\usepackage{hyperref}       
\usepackage{url}            
\usepackage{booktabs}       
\usepackage{amsfonts}       
\usepackage{nicefrac}       
\usepackage{microtype}      
\usepackage{lipsum}
\usepackage{fancyhdr}       
\usepackage{graphicx}       

\usepackage{underscore}
\usepackage{algorithm}
\usepackage[noend]{algpseudocode}

\floatname{algorithm}{Algorithm}

\graphicspath{{media/}}     

\title{TRR360D: A dataset for 360 degree rotated rectangular box table detection
}

\author{
  Wenxing Hu \\
  School of Electronics \& Information Engineering\\
  Shanghai University of Electric Power\\
  Shanghai 200090, China \\
  \texttt{moonstarwork@gmail.com} \\
   \And
  Minglei Tong \\
  School of Electronics \& Information Engineering\\
  Shanghai University of Electric Power\\
  Shanghai 200090, China \\
  \texttt{tongminglei@gmail.com} \\
}

\begin{document}
\maketitle

\begin{abstract}
To address the problem of scarcity and high annotation costs of rotated image table detection datasets, this paper proposes a method for building a rotated image table detection dataset. Based on the ICDAR2019MTD modern table detection dataset, we refer to the annotation format of the DOTA dataset to create the TRR360D rotated table detection dataset. The training set contains 600 rotated images and 977 annotated instances, and the test set contains 240 rotated images and 499 annotated instances. The AP50(T<90) evaluation metric is defined, and this dataset is available for future researchers to study rotated table detection algorithms and promote the development of table detection technology. The TRR360D rotated table detection dataset was created by constraining the starting point and annotation direction, and is publicly available at \url{https://github.com/vansin/TRR360D}.
\end{abstract}


\section{ICDAR2019MTD}
ICDAR2019MTD\cite{Gao2019}  Modern Table Detection dataset is proposed at the Table Detection and Recognition Competition of the 2019 International Conference on Document Analysis and Recognition. It consists of 600 training images and 240 testing images, with 977 annotated table instances in the training set and 449 annotated table instances in the testing set in XML format. The annotation format is shown as \ref{equa:icdar2019}, with the top-left corner of the image as the starting point and other annotation points arranged counterclockwise. This dataset has been widely used in academic research and industrial applications.

\begin{equation}
	x_{1} \ y_{1} \ x_{2} \ y_{2} \ x_{3} \ y_{3} \ x_{4} \ y_{4}
\label{equa:icdar2019}
\end{equation}

\begin{figure}[ht]
    \centering
    \includegraphics[width=1.0\textwidth]{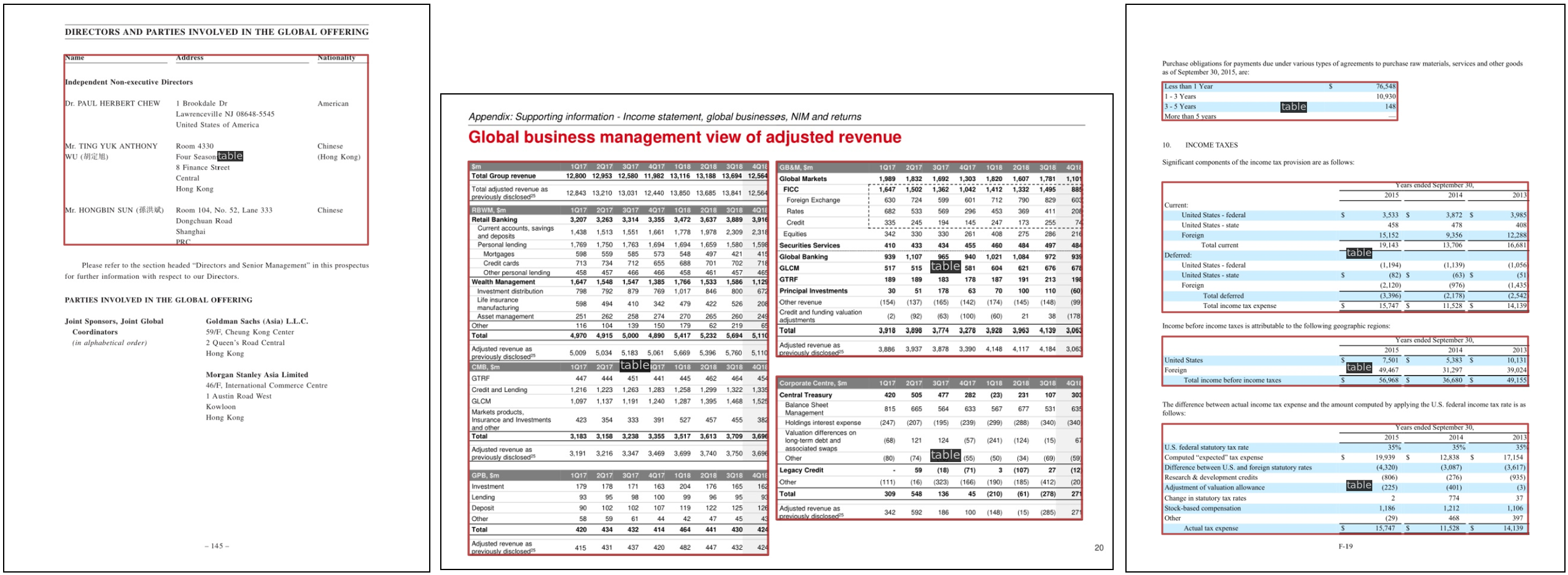}
    \caption{ICDAR2019MTD Visualization}
    \label{fig:orinann}
\end{figure}

One of the limitations of the ICDAR2019MTD dataset is that it only contains horizontally-aligned tables, which cannot be used to train table rotation object detectors. Additionally, the annotation format using four points does not provide semantic information and does not specify the starting point as the top-left corner of the table object. To better utilize the capabilities of the MMRotate\cite{Zhou2022} rotation object detection algorithm, this chapter converts the original ICDAR2019MTD dataset annotations in XML format to DOTA-format text annotations and introduces the constraints and methods for creating TRR360D annotations.

\section{TRR360D Annotation Format}

According to the DOTA\cite{Xia2018} dataset annotation format, a line in the text file corresponding to the annotation of a rotated table instance is shown \ref{equa:TRR360D}. Point A represents the top-left point of the table, and points ABCD are arranged clockwise. Point D represents the detection difficulty of the sample, which is uniformly defined as 0 in TRR360D.

\begin{equation}
	x_{A} \ y_{A} \ x_{B} \ y_{B} \ x_{C} \ y_{C} \ x_{D} \ y_{D} \ table \ D 
 \label{equa:TRR360D}
\end{equation}

In this format, x1, y1, x2, y2, x3, y3, x4, and y4 represent the x and y coordinates of the four points A, B, C, and D, respectively. The last column represents the detection difficulty, which is set to 0 for all samples in TRR360D.

\section{Manually adjusted portion of labels}

\begin{figure}[ht]
    \centering
    \includegraphics[width=0.9\textwidth]{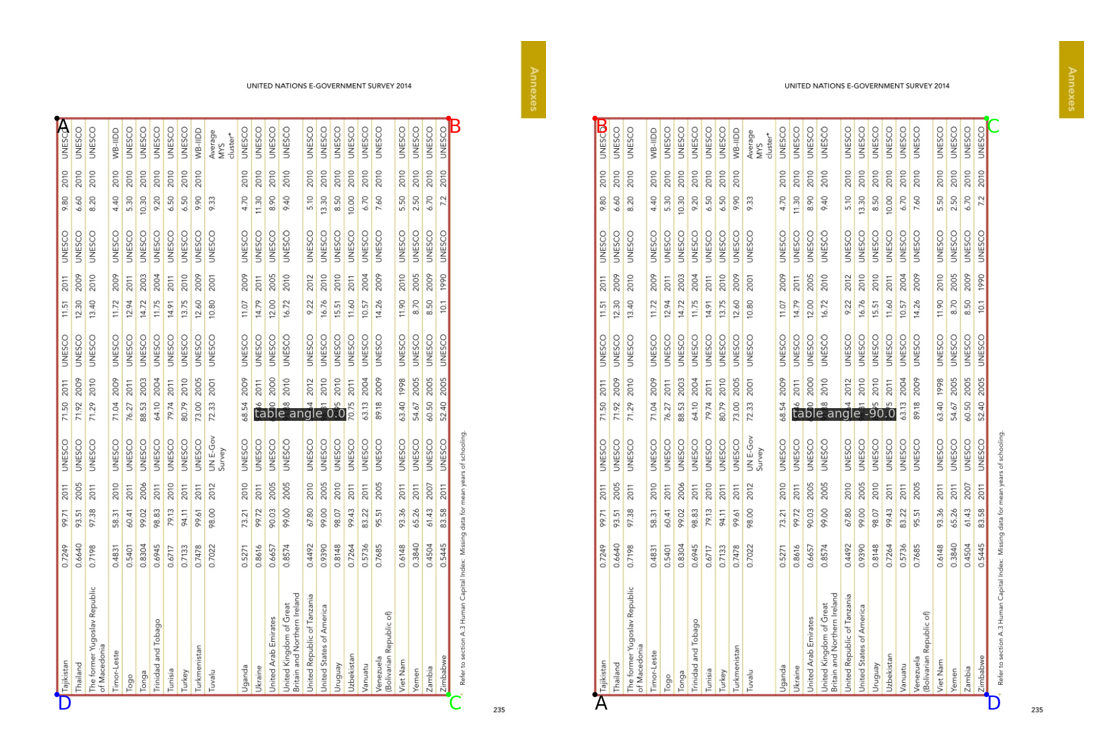}
    \caption{Adjustment of starting point labeling}
    \label{fig:AdjustPoint}
\end{figure}

Figure \ref{fig:AdjustPoint} On the left shows the visualization of the original annotation of image 10497 in the ICDAR2019MTD modern table detection dataset, where the ABCD four-point coordinates are shown as Equation \ref{equa:beforeadjust}. Since point A is not in the upper left corner of the table, the labeling points need to be adjusted so that point A is in the upper left corner of the table and points BCD satisfy the clockwise constraint.

\begin{equation}
	63 \ 119 \ 666 \ 119 \ 666 \ 1006 \ 63 \ 1006 \ table \ 0
 \label{equa:beforeadjust}
\end{equation}

By manually adjusting the coordinates to the order of \ref{equa:afteradjust}, the visualization result is shown in the right image of Figure \ref{fig:AdjustPoint}. At this point, point A is located in the top-left corner of the table, and points BCD satisfy the clockwise constraint.

\begin{equation}
	63 \ 1006 \ 63 \ 119 \ 666 \ 119 \ 666 \ 1006 \ table \ 0
 \label{equa:afteradjust}
\end{equation}

There are some images in the original dataset that do not satisfy the constraints. Therefore, the starting point positions of images 10001, 10037, 10149, 10206, and 10211 in the testing set were manually adjusted. Additionally, the point order of images 10062, 10108, 10187, 10418, 10445, 10497, and 10537 in the training set were manually adjusted to satisfy the constraints of point A at the top-left corner of the table and points ABCD arranged clockwise.

\section{Rotated Dataset Generation}

The aforementioned work has converted the XML files of the ICDAR2019MTD dataset to txt files and manually adjusted the annotations of all samples with anomalous starting points, similar to Figure \ref{fig:AdjustPoint}. In this section, we propose an adaptive boundary rotation mapping method based on the original OpenCV rotation mapping method. This method applies random rotation transformations to the existing images and annotated coordinates, generating the TRR360D rotated table detection dataset.

\subsection{OpenCV Original Rotation}

\begin{figure}
    \centering
    \includegraphics[width=0.9\textwidth]{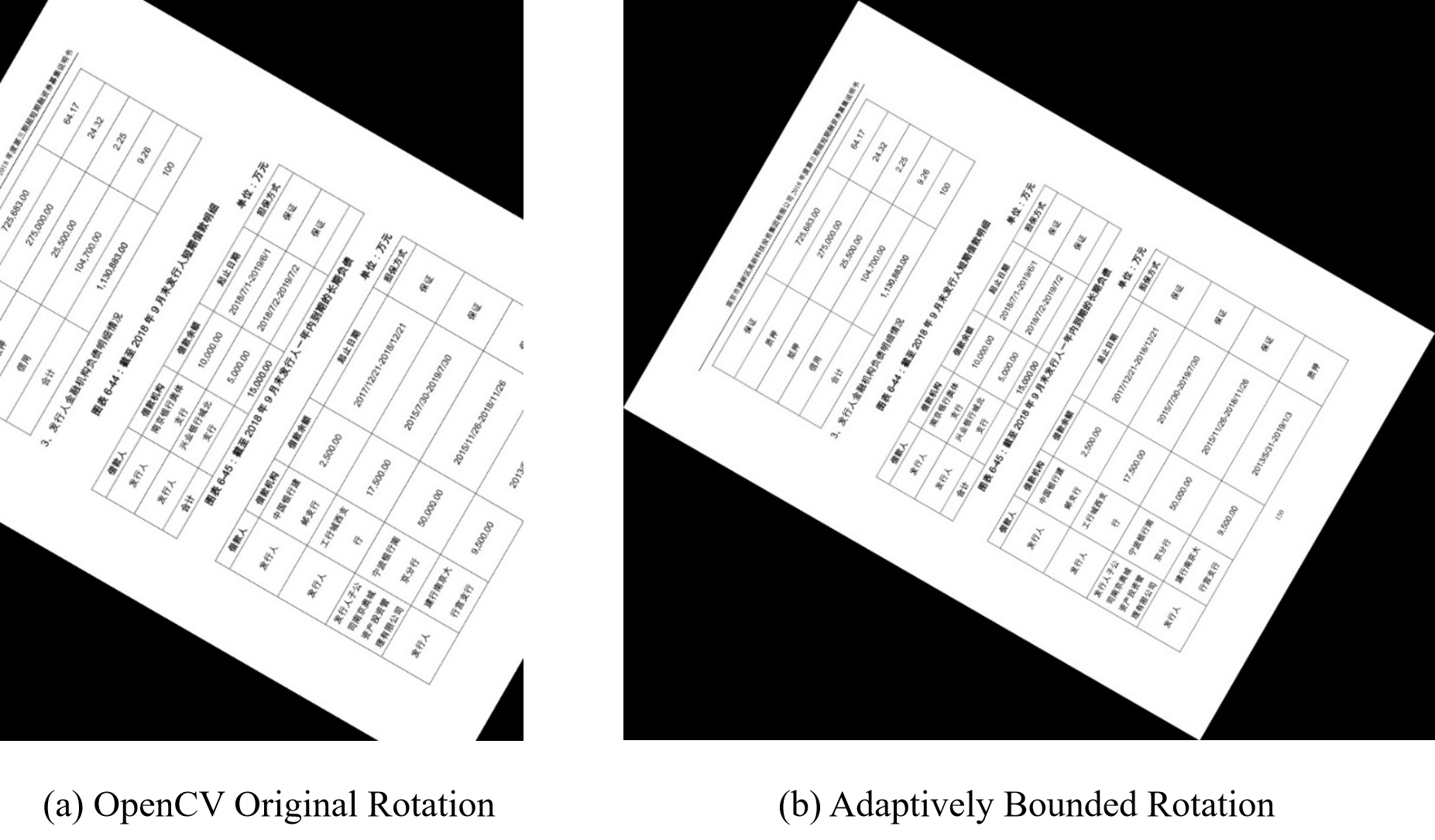}
    \caption{Rotation transformation}
    \label{fig:Rotate}
\end{figure}

Algorithm \ref{algorithm:a1} describes a method that can achieve the default OpenCV rotation, as shown in  Figure \ref{fig:Rotate}(a), by rotating the image and mapping the rotated annotation coordinates. However, the drawback of this method is the loss of the table region.

\begin{algorithm}
\caption{OpenCV Original Rotation}
\begin{algorithmic}[1]
\Require image, $\theta$, points: List[(x,y)]
\Ensure r\_image, r\_points: List[(x,y)]

\State h$\gets$image.height
\State w$\gets$image.weight
\State matrix$\gets$cv2.getRotationMatrix2D(center, -angle, 1)

\State r_image$\gets$cv2.warpAffine(image, matrix, (w, h))
\State pts $\gets$ points.reshape([-1, 2])
\State pts $\gets$ np.hstack([pts, np.ones([len(pts), 1])]).T
\State points $\gets$ matrix@points
\State r_points $\gets$ [[points[0][x],points[1][x]] for x in range(len(points[0]))]

\State \textbf{return} r\_image, r\_points
\end{algorithmic}
\label{algorithm:a1}
\end{algorithm}

\subsection{Adaptively Bounded Rotation}
We propose an adaptive boundary rotation transformation algorithm to overcome the problem of losing table regions when rotating images. The algorithm flow is shown in Algorithm \ref{algorithm:a2}, and the transformation effect is shown in Figure \ref{fig:Rotate}.

\begin{algorithm}[h]
\caption{Adaptively Bounded Rotation}
\begin{algorithmic}[1]
\Require image, $\theta$, points: List[(x,y)]
\Ensure r\_image, r\_points: List[(x,y)]

\State h$\gets$image.height
\State w$\gets$image.weight
\State matrix$\gets$cv2.getRotationMatrix2D(center, -angle, 1)

\State cos = abs(matrix[0, 0]) 
\State sin = abs(matrix[0, 1])
\State new_w = h * sin + w * cos
\State new_h = h * cos + w * sin
\State matrix[0, 2]$\gets$matrix[0, 2]+(new_w - w) * 0.5
\State matrix[1, 2]$\gets$matrix[1, 2]+ (new_h - h) * 0.5
\State r_image$\gets$cv2.warpAffine(image, matrix, (new\_w, new\_h))

\State pts $\gets$ points.reshape([-1, 2])
\State pts $\gets$ np.hstack([pts, np.ones([len(pts), 1])]).T
\State points $\gets$ matrix@points

\State r_points $\gets$ [[points[0][x],points[1][x]] for x in range(len(points[0]))]

\State \textbf{return} r\_image, r\_points
\end{algorithmic}
\label{algorithm:a2}
\end{algorithm}

\section{Evaluation}

\subsection{Rotated IoU}

Let the detected bounding boxes from a deep learning model be denoted as predicted boxes $P$, and the annotated boxes in the dataset are denoted as ground truth boxes $G$.

\begin{figure}
    \centering
    \includegraphics[width=0.3\textwidth]{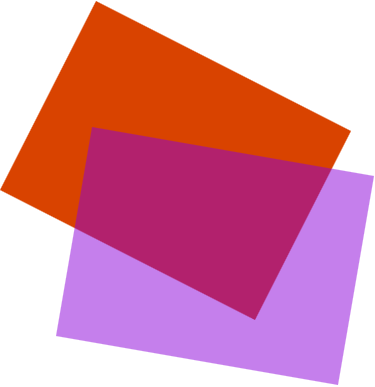}
    \caption{Rotate IoU}
    \label{fig:rotatediou}
\end{figure}

\begin{equation}
	IoU_{PG}=\frac{P \cap G}{P \cup G}
 \label{equa:riou}
\end{equation}


The definitions of the subsequent metrics, including $TP$, $Precision$, $Recall$, $F1Score$, and $AP$, are all related to the $IoU$ between the predicted bounding boxes $P$ and the ground truth boxes $G$.

\subsection{TP FP FN}


$TP$: $True \ Positive$, refers to the predicted boxes $P$ that satisfy the conditions $IoU_{PG}>T_{IoU}$ and $\left|P_{\theta}-G_{\theta}\right|<T_{\theta}$. Here, $T_{IoU}$ is the IoU threshold, which is only counted once for each ground truth box. The larger the threshold, the higher the challenge for locating accuracy. In PASCAL VOC 2007, $T_{IoU}=0.5$. $T_{\theta}$ is the angle threshold, and the smaller the angle, the greater the difficulty.


$FP$: $False Positive$, the number of prediction boxes $P$ satisfying $IoU \leq T_{IoU}$ or $|\theta_P - \theta_G| \geq T_{\theta}$, or the number of redundant prediction boxes detected for the same ground truth box. It is also known as the number of false detections.


$FN$: $False Negative$, the number of ground truth boxes G that were not detected by the model, also known as missed samples.

\subsection{Precision \& Recall}

$Precision$: The ratio of the number of correctly predicted boxes to the total number of predicted boxes.

\begin{equation}
	Precision=\frac{TP}{TP+FP}
 \label{equa:Precision}
\end{equation}

$Recall$: The ratio of correct predictions to the total number of ground truth boxes is the evaluation metric for the dataset.

\begin{equation}
	Recall=\frac{TP}{TP+FN}
 \label{equa:Recall}
\end{equation}

\subsection{RR360 AP50(T<90)}


PR curve: The PR curve is a common performance evaluation metric used to measure the performance of object detection models at different recall and precision levels. The PR curve is plotted by calculating the recall and precision of the object detection model at different confidence threshold levels. The AP value is the area under the PR curve, which is usually computed using the 11-point method for faster computation in practical implementation.


$AP50(T<90)$ refers to the area under the precision-recall (PR) curve at a specific configuration, where the true positive (TP) condition is defined as $IoU_{PG}>0.5$ and $\left|P_{\theta}-G_{\theta}\right|<90$, where $T_{IoU}=0.5$ and $T_{\theta}=90$.

$AP75(T<40)$ refers to the area under the precision-recall (PR) curve at a specific configuration, where the true positive (TP) condition is defined as $IoU_{PG}>0.75$ and $\left|P_{\theta}-G_{\theta}\right|<40$, where $T_{IoU}=0.75$ and $T_{\theta}=40$.




\section{Conclusion}

To address the problem of scarcity and high annotation costs of rotated image table detection datasets, this chapter proposes a method for building a rotated image table detection dataset. Based on the ICDAR2019MTD modern table detection dataset, we refer to the annotation format of the DOTA dataset to create the TRR360D rotated table detection dataset, as shown in Table \ref{tab:ann} The training set contains 600 rotated images and 977 annotated instances, and the test set contains 240 rotated images and 499 annotated instances. The RR360 AP50(T<90) evaluation metric is defined, and this dataset is available for future researchers to study rotated table detection algorithms and promote the development of table detection technology.

\begin{figure}[ht]
    \centering
    \includegraphics[width=1.0\textwidth]{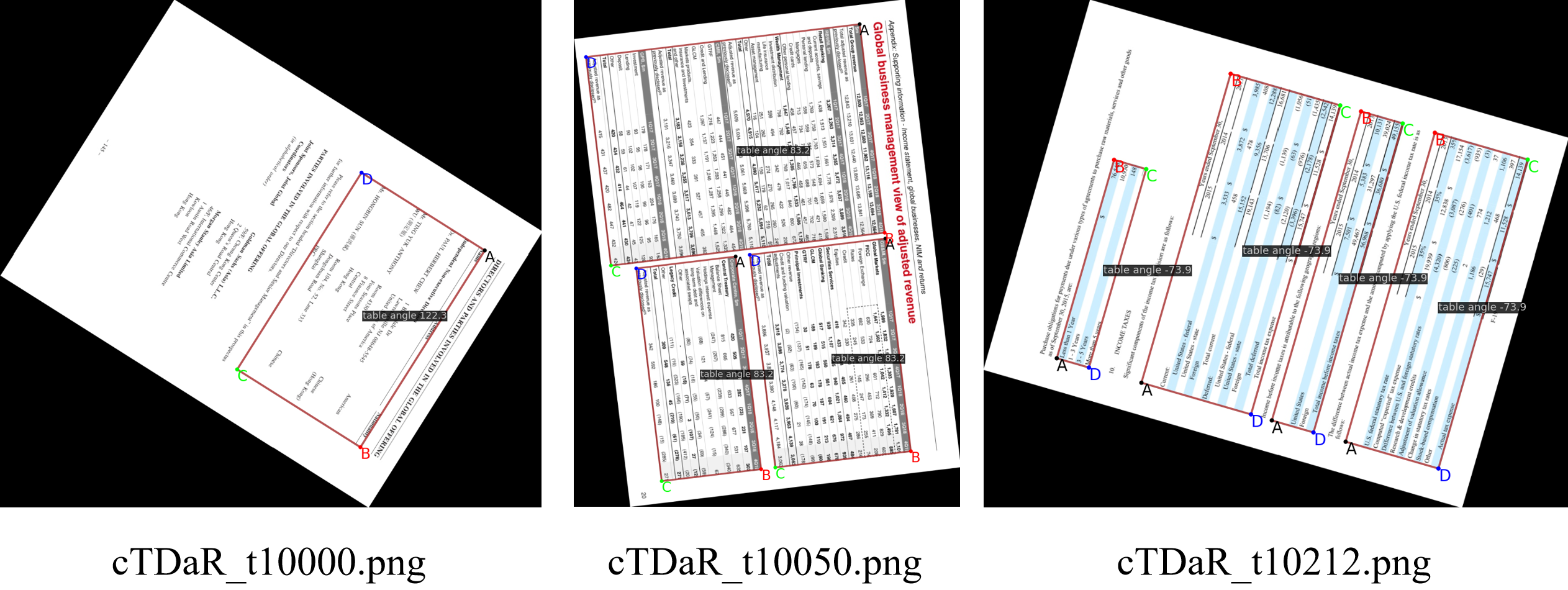}
    \caption{TRR360D Visualization}
    \label{fig:Rotate1}
\end{figure}

\begin{table}
	\caption{TRR360D dataset folders and annotations}
	\centering
	\begin{tabular}{lllll}
		\cmidrule(r){1-5}
		Folder     & Description     & Format & Images & Instances\\
		\midrule
        ann\_test\_hbbox & Horizontal test set annotations & txt & 240 & 449 \\
        ann\_test\_obbox & Rotated test set annotations & txt & 240 & 449 \\
        ann\_train\_hbb & Horizontal training set annotations & txt & 600 & 977 \\
        ann\_train\_obbox & Rotated training set annotations & txt & 600 & 977 \\
		\bottomrule
	\end{tabular}
	\label{tab:ann}
\end{table}

\bibliographystyle{unsrt}  
\bibliography{references}

\end{document}